\documentclass{article}

\usepackage{arxiv}

\usepackage[utf8]{inputenc} 
\usepackage[T1]{fontenc}    
\usepackage{hyperref}       
\usepackage{url}            
\usepackage{booktabs}       
\usepackage{amsfonts}       
\usepackage{algorithm}
\usepackage{algorithmic}
\usepackage{multirow}
\usepackage{nicefrac}       
\usepackage{microtype}      
\usepackage{lipsum}		
\usepackage{graphicx}
\usepackage{natbib}
\usepackage{doi}

\title{Multimodal CLIP Inference for Meta-Few-Shot Image Classification}

\date{} 				

\author{ \href{}{Constance Ferragu}\thanks{Work completed during Bachelor Thesis at InstaDeep and École Polytechnique.} \\
	Princeton University\\
	Princeton, NJ \\
	\texttt{cf7551@princeton.edu} \\
	\And
	{Philomène Chagniot}\thanks{Equal Supervision.} \\
	InstaDeep\\
	Paris, France \\
	\texttt{p.chagniot@instadeep.com} \\
        \And
	{Vincent Coyette}\footnotemark[2] \\
	InstaDeep\\
	Paris, France \\
	\texttt{v.coyette@instadeep.com} \\
}

\renewcommand{\undertitle}{}

\hypersetup{
pdftitle={A template for the arxiv style},
pdfsubject={q-bio.NC, q-bio.QM},
pdfauthor={David S.~Hippocampus, Elias D.~Striatum},
pdfkeywords={First keyword, Second keyword, More},
}

\begin{document}
\maketitle

\begin{abstract}
    In recent literature, few-shot classification has predominantly been defined by the $N$-way $k$-shot meta-learning problem. Models designed for this purpose are usually trained to excel on standard benchmarks following a restricted setup, excluding the use of external data. Given the recent advancements in large language and vision models, a question naturally arises: can these models directly perform well on meta-few-shot learning benchmarks? Multimodal foundation models like CLIP \cite{clip}, which learn a joint \textit{(image, text)} embedding, are of particular interest. Indeed, multimodal training has proven to enhance model robustness, especially regarding ambiguities, a limitation frequently observed in the few-shot setup. This study demonstrates that combining modalities from CLIP's text and image encoders outperforms state-of-the-art meta-few-shot learners on widely adopted benchmarks, all without additional training. Our results confirm the potential and robustness of multimodal foundation models like CLIP and serve as a baseline for existing and future approaches leveraging such models.
\end{abstract}

\section{Introduction}
Generally speaking, few-shot learning (FSL) aims to address the challenge of learning new concepts or tasks with limited labeled data. Initially framed as a classification problem with very little data, the problem definition evolved towards \textit{$N$-way $k$-shot learning}, where the goal is to learn how to discriminate between any $N$ classes given $k$ examples per class, where $k$ is usually small. A wide variety of FSL methods have been developed since, ranging from nearest-neighbor-based algorithms \cite{vinyals2016matching, snell2017prototypical, oreshkin2018tadam, wang2019revisiting} to sophisticated differentiable optimization algorithms \cite{ravi2016optimization, finn2017model, bertinetto2018meta, lee2019meta} and probabilistic approaches \cite{trident, bavardage}. Many of these approaches learn from restricted data sources, excluding the possibility of reusing externally pre-trained backbones.

More recently, simpler methods relying on the fine-tuning of models, pre-trained either in-domain \cite{dhillon2019baseline, chen2019closer, chen2021meta} or externally \cite{bateni2021enhancing, Transductive_CNAPS, hu2022pushing}, have yielded significant improvements over existing few-shot benchmarks. By exploiting prior sources of knowledge, these approaches likely reduce the risk of overfitting, a common concern in the few-shot literature \cite{sun2019meta}.
They rely on fine-tuning to mitigate a potential domain shift, although this process can be complex in a few-shot setup when heavy architectures are used. 

In light of the recent development of powerful language and vision models, a question naturally arises: can these models, trained on large external data sources, directly perform well on few-shot image classification benchmarks? Multimodal foundation models that learn a joint \textit{(image, text)} embedding are of particular interest. Indeed, language and visual semantics have distinct features and structures, which can be more or less discriminative depending on the use case. In addition to learning the language and visual modalities independently, multimodal training can capture relations between the modalities, which has demonstrated improved robustness to ambiguous data \cite{cheng2023robustness, fang2022data}. 

In this work, we leverage CLIP's joint multimodal latent space, learned through contrastive learning, for meta-few-shot image classification, without further training. Our results outperform the latest few-shot learning methods on widely adopted $N$-way $k$-shot image classification benchmarks, confirming the potential and robustness of multimodal foundation models.

\section{Problem Definition: N-way k-shot} \label{problemdef}

The goal of $N$-way $k$-shot learning is to learn how to discriminate between any $N$ classes given $k$ examples per class, where $k$ is usually small. 

The training and evaluation procedures of $N$-way $k$-shot learning differ from classical learning methods. In the classical supervised learning setup, the model learns from a subset of the available examples \textit{for each} class and is evaluated on new samples \textit{from the same set of classes}. Instead, in the $N$-way $k$-shot learning setup, the set of classes seen during training $D_{train}$, validation $D_{valid}$ and testing $D_{test}$ are \textit{disjoint}, and define non-overlapping \textit{meta-train}, \textit{meta-validation} and \textit{meta-test} sets, respectively. An iteration is not defined as a batch $(x,y)$ of inputs and labels, but instead, as an \textit{episode} or \textit{task}, which corresponds to a new $k$-shot classification problem defined on a subset of $N$ classes.

For an \textit{episode} $e$, $N$ classes are sampled from $D_{train}$, $D_{valid}$, or $D_{test}$ for training, validation, and testing, respectively. Then, $k$ labeled examples of each class are sampled to create the \textit{support set}: $S_e = \{(x_i, y_i)\}_{i=1}^{N \times k}$, and $q$ additional examples of each class are sampled to create the \textit{query set}: $Q_e = \{(x_i, y_i)\}_{i=1}^{N \times q}$. During training, the model uses the support set to learn how to differentiate between the $N$ classes. The model gradients are updated based on the loss obtained on the query set. During evaluation, the model uses the support set as input to classify the query set examples. The model performance is measured by the average performance over the query sets from $E$ episodes. 

\section{Related Works}

We focus our review on transfer learning, metric learning, and multimodal learning as the entire domain would be too large to review. 

\subsection{Transfer Learning}
Transfer learning involves leveraging knowledge gained from one task to improve performance on another related task. Given the challenge of data scarcity in FSL, transfer learning becomes a valuable strategy. Early FSL methods primarily focused on adapting typical image classification models for the few-shot setting. For example, \cite{fei2006one} proposes a Bayesian approach that takes advantage of previously learned classes to gain information about a new class from a limited number of images. The objective of these initial methods was to develop models with the ability to generalize and transfer their knowledge to unseen classes. 

\paragraph{Pre-training} Learned representations from pre-training capture valuable information which can be transferred to FSL tasks. These representations can offer strong starting points when little data is available. Recent research has explored the effectiveness of pre-training in few-shot learning. For instance, the authors of \cite{bateni2021enhancing, Transductive_CNAPS} evaluate an ImageNet-pre-trained few-shot model. The P\textgreater M \textgreater F framework proposed by \cite{hu2022pushing} fine-tunes large backbones, pre-trained on several large-scale datasets including ImageNet and WebImageText (WiT), for few-shot tasks. Conversely, recent large vision models are often tested on downstream few-shot classification tasks, to test the quality of learned representations and the model's ability to generalize \cite{clip}. These methods primarily seek to transfer knowledge from external data sources to best infer during their downstream few-shot classification task, leading to competitive results.

Models trained for the $N$-way $k$-shot classification tasks often solely learn from datasets designed for this task, such as CIFAR-FS \cite{bertinetto2018meta}, MiniImageNet (MiniIN) \cite{vinyals2016matching}, or the Meta-Dataset \cite{triantafillou2019meta}. Typically, both the meta-train and meta-test sets are drawn from the same dataset(s), ensuring a controlled experimental setup. However, a controlled data source also limits the development of FSL methods in terms of information source comparisons. In the controlled $N$-way $k$-shot setup, we observe fewer methods that rely on pre-trained backbones, yet, on methods that do, the consensus is that pre-trained backbones increase performance \cite{dhillon2019baseline, chen2019closer, chen2021meta, hu2022pushing}. Although limiting the data source allows for methodical comparisons, leveraging larger data sources could further boost performance.

\subsection{Metric Learning}
One family of FSL methods relies on metric learning, which aims to learn a suitable metric space, in which similar data points are closer together and dissimilar ones are farther apart. These approaches usually consist of three stages: learning the embedding or feature space, constructing class representations, and selecting the similarity metric. 

\paragraph{Feature space} Embedding learning methods can be categorised into task-dependent \cite{oreshkin2018tadam} or task-agnostic \cite{snell2017prototypical, matchingnetworks, liu2018learning} approaches. The former often encounters overfitting issues which can be partially mitigated through data augmentation techniques \cite{zhang2019few}, or task-agnostic pre-training \cite{dhillon2019baseline, chen2019closer, chen2021meta, hu2022pushing}. 

\paragraph{Class representations} Different methods have been used to construct visual class representations. In the 1-shot setup, \cite{koch2015siamese} simply uses the representation of the single image per class in the support set. More generally, when $k>1$, a class prototype can be defined for each class as a weighted combination of the embedded samples from the support set \cite{snell2017prototypical, ren2018meta}, sometimes with learnable parameters \cite{matchingnetworks}. These class prototypes can also be learned end-to-end \cite{ravichandran2019few, yoon2019tapnet}.

\paragraph{Similarity metric} Additionally, a similarity metric is necessary to compare the class representation to the query input. Most methods use a fixed metric, usually the Euclidean distance \cite{snell2017prototypical} or the cosine similarity \cite{matchingnetworks}, while others \cite{sung2018learning, shalam2022selfoptimaltransport} learn a ``relation score", measuring the relative relationship between classes. 

Once the metric space is learned, any image can be classified using the similarity between its embedding and the class representations.

\subsection{Multimodal Learning}
Strong evidence supports the hypothesis that language can help toddlers recognize new visual objects \cite{tod, tod2}. Following this idea, early works investigate coupling visual and textual modalities for zero-shot learning \cite{larochelle2008zero, palatucci2009zero}, through \textit{modality-alignment}. During training, modalities are closely mapped together, thus enabling inference using the similarity between the input image and textual representations of different classes. More recently, some approaches have used contrastive text-image alignment as an additional objective or as a fine-tuning step \cite{afham2022visualsemantic, jiang2023film}.

However, language and visual semantics have distinct features and structures, which can be more or less discriminative, depending on the use case. Indeed, some features are more discriminative textually rather than visually (eg. when comparing a ping pong ball with an egg), and vice versa. This idea was investigated in Adaptive Modality Mixture Mechanism (AM3) \cite{xing2019adaptive}, which explores the combination of visual and textual embeddings in the context of \textit{Prototypical Networks} \cite{snell2017prototypical}. AM3 introduces a mixing network that learns a weighting factor to combine visual and textual embeddings when constructing class prototypes, which allows us to prioritize the most discriminative modality.

\subsection{CLIP}

CLIP \cite{clip}, a vision-language pre-trained model, learns the general relationship between image and text pairs, providing a foundational understanding of textual and visual similarities across a wide array of concepts. CLIP's multi-modality paired with metric learning techniques enable the downstream transfer of this knowledge for few-shot learning tasks. 

CLIP is trained on the WebImageText (WIT) dataset, which contains 400 million \textit{(image, caption)} pairs corresponding to 500,000 distinct text queries on a variety of publicly available internet sources. CLIP is trained in a contrastive manner: given $N$ \textit{(image, text)} pairs, the cosine similarity metric $s$, and the text and image encoders $f_T$ and $f_I$, $N^2$ similarity scores are computed for each possible combination of image $x$ and text $c$:
\begin{equation}
    sim(c,x)=s(f_T(c), f_I(x))
    \label{eq:sim_score}
\end{equation}
The model is trained with the symmetric cross-entropy loss to maximize the similarity between the $N$ matching pairs while minimizing it for the $N^2-N$ others. This allows for the joint training of an image encoder and a text encoder that share the same latent space. 

At inference time, given the similarity matrix containing the images as rows, and the texts as columns, we can classify a given image by applying a softmax to its corresponding row. Conversely, text-to-image classification can be obtained by applying the softmax over the columns. CLIP can easily be used as a zero-shot image classifier, simply by embedding the set of classes and by computing their similarity with test images. It can be noted that CLIP's contrastive learning task resembles the task being learned by few-shot metric learning methods such as matching networks \cite{matchingnetworks} and AM3 \cite{xing2019adaptive}. 

CLIP outperforms existing SOTA models on major zero-shot image classification benchmarks, including ImageNet. In addition, the authors evaluate the quality of the learned representations through a ``linear probe", which consists in training a shallow linear model on top of the frozen image encoder. This benchmark shows competitive results compared to the existing SOTA vision models. This suggests that CLIP holds a significant amount of prior knowledge which can generalize quickly to new distributions. This linear probe is also used as a benchmark in the \textit{non-meta} few-shot classification setup (as opposed to the $N$-way $k$-shot setup), which again shows promising results, although the average few-shot performance degrades in comparison to the zero-shot setup when $k$ is small. This confirms the complexity of transfer learning from a large model on a few-shot dataset.

\subsection{CLIP-based Few-shot Methods} Recent works have leveraged the multimodal power of CLIP for few-shot problems. For instance, CoOp \cite{zhou2022learning} and CoCoOp \cite{zhou2022conditional} propose novel approaches to automate prompt engineering to improve manual prompt tuning, which requires domain expertise and is time-consuming. The pre-trained weights are kept fixed and learnable vectors generate the prompts. With only a few examples, the learned prompts achieve significant improvements over the manual prompts, allowing CLIP to adapt to diverse few-shot tasks. Proto-CLIP \cite{chen2022prototypical} proposes the integration of prototypical representations within a CLIP-like training objective to enhance the discriminative power of CLIP. This leads to a model which can more effectively generalize and adapt to few-shot tasks. Importantly, these works, along with \cite{zhang2021tip}, \cite{song2022clip}, evaluate with a $k$-shot setup, similar to the CLIP linear probe evaluation pipeline, which differs from the $N$-way $k$-shot meta-few-shot setup considered in this work.

\section{Method} 

Our objective is to assess the performance and the robustness of CLIP \textit{without further training} in the \textit{$N$-way $k$-shot setup}. We follow the standard $N$-way $k$-shot evaluation procedure detailed in section \ref{problemdef}. We want to emphasize that this setup is used for the \textit{meta}-few shot classification problem, as opposed to the ``linear probe'' few-shot setup considered in the CLIP paper and subsequent CLIP few-shot literature, which considers a typical classification setup but with fewer examples.  

In the following experiments, we propose three inference methods that exploit different information from the support set: the textual modality, the visual one, or both. We define the evaluation procedure for an episode $e$ for each of the three inference methods. Let $\{c_1, ..., c_N\} \subseteq \mathcal{C}_{test}$ be the sampled classes, $S_e = \{(x_i, y_i)\}_{i=1}^{N \times k}$, be the support set and $Q_e = \{(x_i, y_i)\}_{i=1}^{N \times q}$ be the query set for episode $e$. 

\subsection{Textual Inference}

\paragraph{Inference} We construct a class representation for each class \texttt{$R_{c_i}$} by averaging the embeddings over the prompts. Given $f_T$, the text encoder, and $t \in \mathcal{T}$, the corresponding promtps, we define:
\begin{equation}
    \texttt{$R_{c_i}$}=\frac{1}{|
     \mathcal{T}|} \sum_{t \in \mathcal{T}}f_T\big( t(c_i) \big)
     \label{eq:rc_textual}
\end{equation}
We compute a cosine similarity matrix $S \in \mathbb{R}^{q\times N}$ between the embeddings of the query images $\{f_I(x_i)\}_{i=1}^{N\times q}$ and the class representations $\{R_{c_i}\}_{i=1}^N$. 
For a given image $x_i$, the classification score is obtained by applying the softmax over the $i^{th}$ row of $S$, eventually with a temperature parameter $\tau$ optimized during CLIP's training:
\begin{equation}
  p(y_{i}=c|x_i, \tau) = \frac{\exp\big(s\big(f_I(x_i), \texttt{$R_c$}\big)/\tau\big)}{\sum_{c'\in \{c_1, ..., c_N\}} \exp\big(s\big(f_I(x_i), \texttt{$R_{c'}$}\big)/\tau\big)}
  \label{eq:softmax}
\end{equation}
The class with the highest likelihood is predicted as the classification output for the image in that row. 

\paragraph{Prompt engineering} CLIP was trained to match images to internet captions, which were expressed in ``natural language". In order to resemble this training distribution, we convert the class names into phrases before being fed to the text encoder. The same \textit{prompts} are used for all classes of a given dataset. For example, for a flower classification task, the template “A photo of a \{label\}, a type of flower.” is a sensible option. When multiple templates can fit a given dataset, the class representation is defined as the average of the embeddings obtained with the different prompts. Since all examples from the same class share the same prompt and class label, their text embeddings are identical. Thus, the textual inference leads to identical results when varying $k$, the number of ``shots" per class.

\subsection{Visual Inference}

\paragraph{Inference} The class representation for each class, \texttt{$R_{c_i}$}, is the \textit{centroid} of the embeddings of all images belonging to this class in the support set. Given $S_c^e$, the $k$ support set examples for class $c$ in the episode $e$, and the image encoder $f_I$, the representation of class $c_i$, $R_{c_i}$, is defined as:

\begin{equation}
    R_{c_i} = \frac{1}{k} \sum _{x_i\in S_c^e} f_I(x_i)
    \label{eq:rc_visual}
\end{equation}

The inference is obtained following equation \ref{eq:softmax}, with the adapted $R_c$, defined in equation \ref{eq:rc_visual}. This approach is similar to the one of matching networks \cite{vinyals2016matching} or prototypical networks \cite{snell2017prototypical}. We note that P\textgreater M\textgreater F \cite{hu2022pushing} uses the same inference at evaluation, although in their case, $f_I$ is fine-tuned on the meta-train set.

\subsection{Stacked Inference}

We also propose an inference method which aggregates both visual and textual modalities.
This aggregation can be applied at different levels, either in the embedding space, by combining the visual and textual class representations, or at the score level, by leveraging the distributions obtained in the visual and textual inference methods detailed above. AM3 \cite{xing2019adaptive} follows the first approach, however, they rely on a learned parameter to weigh the relative importance of both types of embeddings. Instead, we choose to focus on aggregating the models at the score level, which allows us to prioritize the predictions based on the relative confidence of the two models.

\paragraph{Inference} To evaluate a query image $x_i$, we compute the visual inference distribution and textual inference distribution defined by equation \ref{eq:softmax}. The stacked inference combines these distributions to obtain a third distribution to use for classification. We experiment with two simple aggregation methods: the class-wise \textit{maximum} or \textit{average} of the two distributions. 

In a case where the visual inference and textual inference yield different classifications, by taking the maximum likelihood, we choose to prioritize the most confident prediction. Alternatively, taking the average allows us to handle cases where the models disagree on their preferred option, but agree on their ``second best" choice, which is more robust to potential outliers. 

One difficulty comes from the potential miscalibration of CLIP, which could be exacerbated with domain shift. Indeed, in a case where one of the textual or visual models is over or underconfident, combining their scores could hinder performance. This issue could be partially mitigated by the fact that both CLIP encoders share the same embedding space, hoping that the eventual domain shift would impact the calibration of both encoders similarly. In addition, recent works \cite{revisiting_calibration} show that the zero-shot CLIP model is well calibrated, both when tested on ImageNet, a (relatively) in-domain dataset, and on synthetic out-of-distribution data \cite{imagenet_c, imagenet_a, imagenet_r}. This motivates the idea described above of ensembling without additional calibration.

\begin{algorithm}[tp]
    \caption{Pseudo-code for visual or textual inference using CLIP. }
    \label{alg:general_inference}
    \textbf{Parameters}: $\mathcal{E}$ the set of episodes, \\ $\mathcal{C}_{e}$ the set of classes in episode $e$, \\
    $\mathcal{S}_e^c$ the support set for episode $e$ with class $c$, \\
    $\mathcal{Q}_{e} = \{(X_{Q_e}, Y_{Q_e})\}_i$ the query set for episode $e$, \\
    $\mathcal{T}$ the set of prompts, \\
    $f_T$ the CLIP text encoder, \\
    $f_I$ the CLIP image encoder,\\
    $s$ a similarity measure, e.g. the cosine similarity,\\
    $\tau$ the temperature.
    \begin{algorithmic}[1] 
        \STATE \texttt{accuracies=[ ]}
        \FOR {$e \in \mathcal{E}$}
        \FOR {$c \in \mathcal{C}_{e}$}
        \STATE \texttt{\textit{\# if textual inference: }}
        \STATE \texttt{$R_c$}=$\frac{1}{|
        \mathcal{T}|} \sum_{t \in \mathcal{T}}f_T\big( t(c) \big)$
        \STATE \texttt{\textit{\# if visual inference: }}
        \STATE  \texttt{$R_c$}=$\frac{1}{k} \sum_{x_i \in \mathcal{S}_e^c}f_I\big( x_i \big)$
        \ENDFOR
        \STATE $S = \{ s(f_I(x_i), R_j)\}_{i, j}$ \texttt{\textit{\#image-class similarity matrix}}
        \STATE \texttt{probs} = $S$.\texttt{softmax(axis=1, temp=$\tau$)}
        \STATE $\texttt{pred}$ = $\texttt{probs.argmax(axis=1)}$
        \STATE $\texttt{accuracies.append} \big (\frac{1}{|Q_e|} \sum (\texttt{pred}==Y_{Q_e}) \big)$
        \ENDFOR
        \STATE \textbf{return: } \texttt{accuracies.mean()}
    \end{algorithmic}
\end{algorithm}

\section{Experiments} 
We evaluate our approach following two distinct setups presented below. The experiments were run on 8 CPUs and 1 Tesla-V100-SXM3-32GB GPU. 

\paragraph{Fixed N-way k-shot} In classical benchmarks, $N$ and $k$ are fixed both during training and evaluation. The most common benchmarks are the 5-way 1-shot and 5-way 5-shot setups on CIFAR-FS \cite{bertinetto2018meta} and MiniIN \cite{vinyals2016matching}. CIFAR-FS is a random subset from CIFAR-100 \cite{krizhevsky2009learning}, while MiniIN was randomly sampled from 100 classes of ImageNet. The meta-train, validation, and test splits include 64, 16, and 20 classes, respectively, for both datasets. Following the evaluation setup in \cite{bertinetto2018meta}, we evaluated our approach over $E=10,000$ episodes with a query set of $q = 15$ images per class. 

\paragraph{Varied N-way k-shot} The Meta-Dataset \cite{triantafillou2019meta} introduced a new setting which varies $N$ and $k$ to obtain a more realistic and adaptable setting. The Meta-Dataset is a collection of 10 image classification datasets, including general datasets, such as Image-Net, as well as domain specific ones, such as VGG Flower or GTSRB. For each episode, $N$ is sampled uniformly between $(5, min(50, $ \# classes$))$ and $k$ can vary between 1 and 100. We follow the evaluation setup detailed in \cite{triantafillou2019meta} and evaluate over $E=600$ episodes with a query set of $q=15$ images per class. 

\paragraph{Model} Multiple versions of CLIP exist with different ResNet and ViT image encoder architectures. We present our results with ViT-L/14, the best available model. We also report results with ViT-B/16 in Appendix D for a fairer comparison against P\textgreater M \textgreater F-CLIP which uses this backbone. 

\begin{table}[!ht]
    \footnotesize
    \centering
    \begin{tabular}{c | c  c | c  c } 
    \multirow{2}{4em}{}  & 
    \multicolumn{2}{|c|}{CIFAR-FS} & \multicolumn{2}{|c}{MiniImageNet} \\ 
    & \textbf{1-shot, 5-way} & \textbf{5-shot, 5-way} &\textbf{1-shot, 5-way} & \textbf{5-shot, 5-way}\\
          \toprule
        ProtoNet \cite{snell2017prototypical} &- & - & $49.42 \pm 0.78 $   & $68.20 \pm 0.66$    \\
        PTMAP \cite{shalam2022selfoptimaltransport} & 89.94  & 95.80 & 85.59  & 91.34  \\
         BAVARDAGE \cite{bavardage} &  $87.35 \pm 0.23$  &  $90.63 \pm 0.16$  & $84.80 \pm 0.25$  & $91.65 \pm 0.10$ \\
        TRIDENT \cite{trident} & - & - & $86.11 \pm 0.59$  & $95.95 \pm 0.28$  \\
        PMF \cite{hu2022pushing}  & $84.3$    & $92.2$   & $95.3^?$    & $98.4^?$   \\
        \hline
         \textbf{CLIP Visual} & $69.4 \pm 0.2^* $ &   $96.52 \pm 0.06^* $ & $75.05 \pm 0.23 $  &  $96.65 \pm 0.06 $   \\
        \textbf{CLIP Textual} & $\textbf{98.48}  \pm 0.04^* $ &  $\textbf{98.48}  \pm 0.04^* $ &$ \textbf{99.38}  \pm  0.02 $  &$ \textbf{99.38}  \pm  0.02 $   \\
         \textbf{CLIP Stacked-Max} & $ 93.56 \pm 0.05 ^* $ &  $98.41  \pm 0.08  ^* $ &$ 97.34 \pm  0.11 $  &$ 98.89 \pm 0.06$   \\
        \textbf{CLIP Stacked-Avg} & $ 94.21 \pm 0.12 ^* $ &  $98.47 \pm 0.04 ^* $ &$  97.63  \pm 0.08 $  &$  98.88 \pm  0.09 $   \\
        \bottomrule
    \end{tabular}
    \vspace{3mm}
    \caption{\small Fixed $N$-way $k$ shot:  Benchmark against ProtoNet, and the most recent SOTA models.}\label{fig:vis-fixed}
  \end{table}
  \begin{table}[!ht]
  \footnotesize
  \resizebox{\textwidth}{!}{
  \centering 
  \begin{tabular}{c|c c c c c}
        \textbf{Method} & \textbf{ImageNet } & \textbf{Omniglot } & \textbf{Aircraft } & \textbf{CUB} & \textbf{Textures }\\
        \toprule
        ProtNet, IN \cite{snell2017prototypical} & $50.50\pm1.08$ & $59.98\pm1.35$ & $53.10\pm1.00$ & $68.79\pm1.01 $ & $	66.56\pm0.83$ \\
        ProtNet, ID \cite{snell2017prototypical} & $44.50\pm1.05 $ & $79.56\pm1.12 $ & $71.14\pm0.86$ & $67.01\pm1.02 $ & $65.18\pm0.84$ \\
        \hline
        MD Best, IN \cite{CTX, TSA} & $63.73\pm0.99$ & $82.58\pm1.11$ & $80.13\pm1.01$ & $83.39\pm0.80$ & $79.61\pm0.68$\\
        MD Best, ID \cite{TSA, TriM, Transductive_CNAPS} & $57.35 \pm 1.05$ & $\textbf{94.96} \pm 0.38$ & $89.33 \pm 0.44 $ & $81.42±\pm 0.74$ & $76.74\pm 0.72$ \\
        \hline 
        PMF, IN \cite{hu2022pushing} & $76.69^?$ & $81.42$ & 80.33 & 84.38 & 86.87 \\
        PMF, ID \cite{hu2022pushing} & $77.02^?$ & $91.76$ & $\textbf{89.73} $ & $92.94$ & $86.94$ \\
        \hline
        \textbf{CLIP Visual} & $81.52\pm0.7$ & $68.96\pm1.2^?$ & $76.18\pm0.81$ &  $\textbf{96.74}\pm0.33^?$& $85.59\pm0.54$ \\
        \textbf{CLIP Textual} & $84.37\pm0.57$ & $6.48\pm1.23^?$ & $70.47\pm0.78$ & $91.96\pm0.38^?$ & $82.18\pm0.44$    \\
        \textbf{CLIP Stacked-Max} & $\textbf{86.04}\pm0.52$ &$ 68.75\pm1.23^?$ & $78.04\pm0.75$ &  $96.36\pm0.24^?$ & $87.64\pm0.45$ \\
        \textbf{CLIP Stacked-Avg} & $86.03\pm0.55$ & $68.67\pm1.23^?$ & $78.58\pm0.73$ &  $96.37\pm0.25^?$ & $\textbf{87.9}\pm0.45$  \\
        \bottomrule
    \end{tabular}}
    \end{table}
    \vspace{-0.2cm} 
    \begin{table}[!ht]
    \resizebox{\textwidth}{!}{
    \centering
    \begin{tabular}{c|c c c c c}
        \textbf{Method} & \textbf{QuickDraw} & \textbf{Fungi} & \textbf{VGG Flower} & \textbf{Traffic Signs} & \textbf{MSCOCO}\\
        \toprule
        ProtNet, IN \cite{snell2017prototypical} & $48.96\pm1.08 $ & $39.71\pm1.11$ & $85.27\pm0.77$ & $	47.12\pm1.10$ & $	41.00\pm1.10$ \\
        ProtNet, ID \cite{snell2017prototypical} & $64.88\pm0.89$ & $	40.26\pm1.13$ & $86.85\pm0.71$ & $	46.48\pm1.00 $ & $39.87\pm1.06$\\
        \hline
        MD Best, IN \cite{CTX, TSA} & $71.03\pm0.84$ & $51.38\pm1.17$ & $94.05\pm0.45$ & $	81.71\pm0.95$ & $61.67\pm0.95 $\\
        MD Best, ID \cite{TSA, TriM, Transductive_CNAPS} & $\textbf{82.01}\pm0.57$ & $67.40\pm0.99 $ & $92.18±\pm0.52 $ & $83.55\pm0.90$ & $55.75\pm1.06$\\
        \hline 
        PMF, IN \cite{hu2022pushing} & 75.43 & 55.93 & 95.14 & 89.68 & 65.01 \\
        PMF, ID \cite{hu2022pushing} & $80.2 $ & $\textbf{78.28}$ & $95.79$ & $\textbf{89.86}$ & $64.97$\\
        \hline
        \textbf{CLIP Visual}  & $65.09\pm0.83^?$ & $59.38\pm1.12^?$ & $\textbf{98.44}\pm0.52$ & $66.99\pm0.84$ &  $55.1\pm0.75^?$\\
        \textbf{CLIP Textual} & $44.28\pm0.77^?$ & $23.21\pm0.96^?$ & $89.41\pm0.46$ &  $46.28\pm0.81$ & $70.42\pm0.52^?$   \\
        \textbf{CLIP Stacked-Max}  & $59.09\pm0.83^?$ & $59\pm 1.07^?$ &  $98.35\pm0.13$ & $68.41\pm0.80$ &  $70.46\pm0.61^?$\\
        \textbf{CLIP Stacked-Avg}  & $61.44\pm0.79^?$ & $59.67\pm 1.03^?$ &  $98.28\pm0.14$& $68.81\pm0.79$ &  $\textbf{70.6}\pm0.56^?$\\
        \bottomrule
    \end{tabular}
    }
    \vspace{3mm}
    \caption{ \small Varied $N$-way $k$-shot: Benchmark on the Meta-Dataset (MD) against \textit{MD-Best} (best results \textit{per dataset} according to the MD leaderboard\textsuperscript{\ref{leaderboardlink}}), \textit{P\textgreater M\textgreater F} and ProtNet. \textit{IN}: Models only trained on MD ImageNet. \textit{ID}: Models trained on 8 MD training datasets. We provide an indication of potential data overlap: $^*$ when the impact of data overlap on performance is statistically significant, and $^?$ if we do not have access to the information.}
    \label{fig:vis-variable}
\end{table}

\paragraph{} We present our results in Tables \ref{fig:vis-fixed} and \ref{fig:vis-variable}. We emphasize that comparing these results is not straightforward since the training procedures differ. In Figure \ref{fig:vis-fixed}, ProtoNet \cite{snell2017prototypical}, PTMAP \cite{shalam2022selfoptimaltransport}, BAVARDAGE \cite{bavardage} and TRIDENT \cite{trident} do not rely on external pre-training, while ours and P\textgreater M\textgreater F \cite{hu2022pushing} do. In Figure \ref{fig:vis-variable}, ProtoNet and MD-Best are trained ``in-domain", i.e. on ImageNet or the meta-train set of the Meta-Dataset, while our approach uses external pre-training, and P\textgreater M\textgreater F relies on both. Thus, the results cannot be used as a definitive measure of the quality of the meta-learning algorithms. However, they still provide interesting insights about the potential and the robustness of foundation models and allow us to compare models with similar training and evaluation setups.

Note that, MD-Best records the best performance \textit{per dataset, among all algorithms} reported in the Meta-Dataset leaderboard\footnote{\label{leaderboardlink}See https://github.com/google-research/meta-dataset.}. For ImageNet-only training, MD-Best combines CTX \cite{CTX} and TSA \cite{TSA}, while for in-domain training, MD-Best includes TSA \cite{TSA}, TriM \cite{TriM} and Transductive-CNAPS \cite{Transductive_CNAPS}. 

\subsection{CIFAR-FS and MiniImageNet} 
Our textual inference outperforms or matches P\textgreater M\textgreater F, BAVARDAGE, PTMAP, and TRIDENT SOTA benchmarks on both CIFAR-FS and MiniIN. 
The visual inference also outperforms or matches the SOTA benchmarks in the 5-shot setting, while the model only outperforms ProtoNet in the 1-shot setting. This discrepancy could be explained by the higher sensitivity to domain shift when $k$ is small since the variance in the “representativity” of support set samples is more important. Finally, the stacked inference yields comparable results to the textual inference, suggesting that the stacked method is able to prioritize the more discriminative modality during inference.

\subsection{Meta-Dataset} 

\paragraph{Textual inference} The textual inference results demonstrate SOTA performance on ImageNet, MSCOCO and CUB. On the other datasets, the degraded performance is likely exacerbated with domain shift, or due to the lack of expressivity of the class names. For example, the labels of Fungi are scientific fungi species, which are probably largely out-of-distribution with respect to CLIP's training data. The worst performance is obtained on Omniglot, where labels are of the form \texttt{<alphabet> character <number>} (eg: Tagalog character 1), where the \texttt{<number>} is arbitrary. Thus, the textual embeddings fail to provide sufficient information to distinguish characters within the same alphabet. 

\paragraph{Visual inference} The visual inference performes strongly overall, outperforming certain SOTA results. Interestingly, in cases where the textual inference did not perform well (e.g., Omniglot, Fungi, and GTSRB), the visual inference exhibits better results, and vice-versa (e.g., ImageNet and MSCOCO). The overall performance improvement compared to ProtoNet confirms the quality of CLIP's pre-training since both methods follow the same inference procedure at test time. 

\paragraph{Stacked inference} The best performance is observed with the maximum and average stacking inference methods. A significant advantage of this approach is its robustness to any few-shot problem, regardless of the most discriminative modality. For instance, when considering Omniglot, CUB, Fungi, Flower, and MS-COCO, there is a clear performance gap between the two unimodal models. Still, the stacked inference reaches a performance similar to the best model. For other datasets (ImageNet, Aircraft, Textures, and Traffic), the stacked method outperforms the best of the two unimodal methods. These results suggest that the model confidence sensibly weighs the two modalities.

\section{Discussion} 

\subsection{Meta-learning Benchmark} We emphasize that our benchmarks cannot be strictly considered as a meta few-shot learning method due to two reasons. First, our model is not trained in the ``meta-learning'' setup, in fact, our method does not make use of the meta-train set. In addition, the use of external data for pre-training can be considered a violation of the restricted FSL problem, which complicates the comparison of methods. Still, P\textgreater M\textgreater F is pre-trained on different external datasets, and the Meta-Dataset benchmark allows for pre-training on ImageNet (``IN-only"), or a subset of each dataset included in the Meta-Dataset collection (``In-domain").

\paragraph{Comparison against P\textgreater M\textgreater F} 
We focus our comparisons against P\textgreater M\textgreater F models, as these also rely on externally pre-trained backbones, although P\textgreater M\textgreater F additionally trains on ``in-domain'' or ImageNet data. Against CIFAR-FS and MiniIN, our stacked inference outperforms P\textgreater M\textgreater F models. As for the Meta-Dataset, we match or outperform P\textgreater M\textgreater F models on 5 datasets, and on 6 datasets when comparing against P\textgreater M\textgreater F - IN. These results confirm the potential of multimodal foundation models like CLIP for learning-free baselines. Additionally, we observe similar patterns when comparing against P\textgreater M\textgreater F \textit{with the same} CLIP-B/16 backbone (see Appendix A).

\subsection{Data overlap} CLIP was trained on internet data, potentially including a subset of our benchmarks. Since the data is not publicly available, a comprehensive data leakage analysis is not feasible. However, we can still rely on information related to the construction of the dataset, and \cite{clip}'s extended overlap analysis. 

For each dataset, CLIP's authors estimate the percentage of overlap with a ``duplicate detector". They measure the potential gap in performance when evaluating zero-shot CLIP on the corrupted dataset, against its ``clean" version without duplicates. No significant overlap was detected between WiT and ImageNet, allowing for a zero-shot evaluation of CLIP on ImageNet. The same applies to MiniIN and the MD-ImageNet, which both stem from ImageNet. However, the authors' analysis reveals a 5\% overlap that improves CLIP's performance with CIFAR-100 (from which CIFAR-FS is sampled). They do not detect any overlap that significantly improves the performance of CLIP on FGVC-Aircraft, DescribableTextures, VGG Flower, and GTSRB. Generally speaking, the probability of observing the same \textit{(image, text)} pair between CLIP’s training data and the Meta-Dataset is relatively low if datasets are not leaked in their raw version on the internet. Additionally, with the recent advances in self-supervised learning, observing the same image during pre-training and at inference could constitute a relatively realistic setup, given the vast amount of data available on the internet.

We acknowledge that this is an important factor to take into account when making comparisons to models solely trained on the Meta-Dataset. That is why we compare against P\textgreater M\textgreater F, and additionally provide a dataset overlap indication in Figure \ref{fig:vis-fixed} and \ref{fig:vis-variable}. Among the 12 datasets considered, 6 have a potential overlap. There does not seem to be a noticeable difference in performance depending on the overlap. Indeed, our methods outperform the existing SOTA on 4 datasets without overlap (MiniIN, MD-ImageNet, Textures, and Flowers), and 3 datasets with potential overlap (CIFAR-FS, CUB, and MSCOCO).

\subsection{Domain shift} When using large amounts of external data, we probably trade off overfitting against domain shift. Although CLIP was shown to generalize well on natural data, its performance is still quite variable depending on the dataset. In particular, CLIP performs relatively poorly on ``simple tasks" (eg. MNIST \cite{mnist}). Among potential domain shifts, classification problems with semantically poor labels lead to degraded results when leveraging the text encoder, as observed on Omniglot. In this work, we did not consider any domain shift adaptation, apart from (manually) adapting the prompts, which allows to better match the distribution of the texts from the CLIP training data.

\subsection{Possible extensions and future work.} The main objective of this work is to present a benchmark for few-shot learning by inferring from a foundational model without any additional learning or fine-tuning. We point out a couple of ideas from the many possible extensions. 

First, this benchmark was built using CLIP, while fine-tuned versions of CLIP trained with additional (self-)supervised objectives could also be leveraged \cite{mu2021slip, li2022declip, chen2022prototypical}. Alternative multimodal backbones could also be used, including ALIGN \cite{align}, or more recent multi-modal LLMs \cite{lu2024mllm}. More generally, any vision model could be considered for visual inference, though they do not allow cross-modality comparisons which are required for the textual and stacked inferences.

Another natural extension would be to fine-tune the backbone on the few-shot benchmark, which can be done in several ways. The image and text encoder can be fine-tuned, though the relative size of the network compared to the amount of data might require sophisticated approaches \cite{robust_fine_tuning}. The combination of the textual and visual inference could also be improved, either at the embedding level, following the idea of \cite{xing2019adaptive}, or at the score level, by learning a proper stacking or ensembling model. In addition, generic multi-class calibration methods could be used \cite{on_calibration_of_modern}, as well as methods adapted to the zero-shot \cite{levine2023enabling} or the few-shot \cite{wang2023fewshot} setups, which could help mitigate a potential domain shift.

\section{Conclusion}
In this work, we investigate the use of CLIP, a multimodal foundation model pre-trained on external data, in the few-shot setup. CLIP allows to infer on any set of labels, and can thus serve as a baseline without further fine-tuning or adaptation. We compare different types of inference, based on the textual representation of a given class, the visual one, or the combination of both. We show that inferring from CLIP without additional training can outperform state-of-the-art few-shot learners. Our results confirm the potential of large externally pre-trained foundation models and additionally serve as a baseline for existing and future approaches leveraging such models.

\newpage

\bibliographystyle{unsrtnat}

\newpage 
\appendix
\section{Additional Results against P\textgreater M\textgreater F}
We additionally report results with the ViT-B/16 model for a fairer comparison against P\textgreater M \textgreater F-CLIP, which uses this backbone. P\textgreater M\textgreater F-CLIP is based on the same pre-training dataset (WIT) as our approach, thus it constitutes a fair comparison. However, we do not fine-tune on ImageNet or any in-domain datasets, thus our approach is less data-intensive, in addition to not requiring any fine-tuning. Evaluating against P\textgreater M\textgreater F can be considered as a good baseline.

We observe a similar trend as when using the ViT-L/14 model. Indeed, even when comparing our best model (with average stacking) against P\textgreater M\textgreater F \textit{with the same} CLIP-B/16 backbone (P\textgreater M\textgreater F-CLIP B/16), we again improve on the same 5 datasets, and on 8 datasets if considering P\textgreater M\textgreater F-CLIP-ImageNet only.

\begin{table}[!ht]
  \caption{Additional Varied $N$-way $k$-shot: Benchmark against \textit{P\textgreater M\textgreater F}-CLIP with the B/16 backbone.}
    \label{fig:appendix}
    \resizebox{\textwidth}{!}{
    \centering
    \begin{tabular}{c|c c c c c }
        \textbf{Method} & \textbf{ImageNet} & \textbf{Omniglot} & \textbf{Aircraft} & \textbf{CUB} & \textbf{Textures}\\
        \hline
        PMF-CLIP B/16, IN & 76.03 & $59^?$ & 65.75 & $90.2^?$ & 83.08 \\
        PMF-CLIP B/16, ID & 74.76 & $92.26^?$ & 91.42 & $93.55^?$ & 80.97\\
        \hline
        \textbf{CLIP-B/16 Avg} & $81.23\pm0.41$ & $65.06\pm1.31^?$ & $70.21\pm0.78$ & $94.83\pm0.29^?$ & $84.33\pm0.51$  \\
        \textbf{CLIP-L/14 Avg} & $86.03\pm0.55$ & $68.67\pm1.23^?$ & $78.58\pm0.73$ &  $96.37\pm0.25^?$ & $87.9\pm0.45$  \\
        \hline 
    \end{tabular}
    }
\end{table}
\begin{table}[!ht]
    \resizebox{\textwidth}{!}{
    \centering
    \begin{tabular}{c|c c c c c }
        \textbf{Method} & \textbf{QuickDraw} & \textbf{Fungi} & \textbf{VGG Flower} & \textbf{Traffic Signs} & \textbf{MSCOCO}\\
        \hline
        PMF CLIP B/16, IN &  $65.45^?$ & $53.2^?$ &96.35 & 58.65 & $61.2^?$\\
        PMF CLIP B/16, ID & $80.8^?$ & $79.13^?$ & 95.64 & 54.52 & $56.8^?$\\
        \hline
        \textbf{CLIP-B/16 Avg}  & $47.07\pm0.98^?$ & $54.25\pm1.03^?$ & $97.04\pm0.20$ &  $52.16\pm0.93$ & $67.04\pm0.66^?$\\
        \textbf{CLIP-L/14 Avg}  & $61.44\pm0.79^?$ & $59.67\pm 1.03^?$ &  $98.28\pm0.14$& $68.81\pm0.79$ &  $70.6\pm0.56^?$\\
        \hline
    \end{tabular}
    }
\end{table}

\end{document}